%% file: main.tex
\documentclass[sigconf]{acmart}

\usepackage{color, soul}
\usepackage{booktabs} 
\usepackage{graphicx}
\usepackage{caption}
\usepackage{subcaption}
\usepackage[labelfont=bf,textfont=md]{caption}
\usepackage{balance}

\def\BibTeX{{\rm B\kern-.05em{\sc i\kern-.025em b}\kern-.08emT\kern-.1667em\lower.7ex\hbox{E}\kern-.125emX}}

\begin{document}


\title{Are Accelerometers for\\ Activity Recognition a Dead-end?}

\author{%
  Catherine Tong$^{\dagger}$, Shyam A.~Tailor$^{\dagger}$, Nicholas D.~Lane$^{\dagger \diamond}$  
  }

   \affiliation{
 	\institution{$^{\dagger}$University of Oxford\hspace{+0.5cm}    $^{\diamond}$Samsung AI \vspace{0.58cm}} 
 }

\renewcommand\footnotetextcopyrightpermission[1]{} 
\renewcommand{\authors}{Catherine Tong, Shyam A.~Tailor, Nicholas D.~Lane}

\pagestyle{plain} 

\begin{abstract}
Accelerometer-based (and by extension other inertial sensors) research for Human Activity Recognition (HAR) is a dead-end. This sensor does not offer enough information for us to progress in the core domain of HAR---to recognize everyday activities from sensor data. Despite continued and prolonged efforts in improving feature engineering and machine learning models, the activities that we can recognize reliably have only expanded slightly and many of the same flaws of early models are still present today. Instead of relying on acceleration data, we should instead consider modalities with much richer information---a logical choice are images. With the rapid advance in image sensing hardware and modelling techniques, we believe that a widespread adoption of image sensors will open many opportunities for accurate and robust inference across a wide spectrum of human activities.

In this paper, we make the case for imagers in place of accelerometers as the default sensor for human activity recognition. Our review of past works has led to the observation that progress in HAR had stalled, caused by our reliance on accelerometers. We further argue for the suitability of images for activity recognition by illustrating their richness of information and the marked progress in computer vision. Through a feasibility analysis, we find that deploying imagers and CNNs on device poses no substantial burden on modern mobile hardware. Overall, our work highlights the need to move away from accelerometers and calls for further exploration of using imagers for activity recognition.\footnote{A version of this paper was accepted at The 21st International Workshop on Mobile Computing Systems and Applications (HotMobile) 2020.}
\end{abstract}





\maketitle

\section{Introduction}

For a long time, the default sensors for Human Activity Recognition (HAR) have been accelerometers---low-cost, low-power and compact sensors which provide motion-related information. In the past decades, waves of accelerometer-based (and by extension other inertial sensors) research have enabled HAR studies ranging from the classification of common activities to the objective assessment of their quality, as seen in exciting applications such as skill \cite{khan2015beyond} and disease rehabilitation \cite{patel2012review} assessment at scale. However, despite these success stories, we believe that accelerometers can lead us no further in achieving the primary task at the very core of HAR---to recognize activities reliably and robustly. 

The reliable recognition of activities, first and foremost, requires data to be collected from information-rich and energy-efficient sensors. In 2004, Bao and Intille \cite{Bao2004ActivityRF} published a landmark study to recognize daily activities from acceleration data. 15 years on, we have not moved far from identifying activities much more complex than `walking' or `running' with mobile sensors \cite{wang2019deep}. This slow process is not due to a lack of data (accelerometers are present in almost all smart devices), nor a lack of feature engineering and algorithmic innovation (a great variety of accelerometer-based HAR works exists, including deep learning solutions), but the quality of the sensor data itself. 

It is time to rethink this default HAR sensor and move towards a modality with richer information, in order to identify more activities more robustly. While accelerometers capture motion-related information useful for distinguishing elementary activities, a significant portion of our activities are not characterised by their motions but by their precise context, e.g. social setting or objects of interaction. Recognising these activities unobtrusively is of great interest to many research communities such as psychology and health sciences but currently we are unable to do this with acceleration data. Moving forward, can we rely on such a sensor which inherently lacks the dimension to capture non-motion-based information? Through inspecting years of progress in accelerometer-based HAR research, our answer is no. Our over-reliance on accelerometer-based data is contributing to a bottleneck in inference performance, a constrained list of recognisable activity classes and alarming confusion errors. To overcome this bottleneck, we should instead adopt a sensing modality with much richer information---the most natural choice are \emph{images}.

Against the backdrop of a rapid sophistication of both learning algorithms and sensor hardware, imagers---low-form factor visual sensors---stand as a strong candidate to replace accelerometers as the default sensor in a new wave of HAR research. Notably, the idea that images are information-rich is not new: both stationary and wearable cameras have been used in numerous studies for visual-based action recognition \cite{WEINLAND2011224}, and for providing labelable ground truth and additional contexts \cite{doherty2013using}; The superiority of cameras compared to other sensors, including accelerometers, for high-level activity recognition has also been argued empirically \cite{maekawa}. Critically, previous barriers to the adoption of imagers have now been overcome by substantial advancement seen in recent years: on one hand, powerful and efficient deep learning algorithms have been developed for image processing \cite{sze2017efficient, han2015deep, lane2017squeezing}; on the other hand, there are now miniature image sensors with low energy requirement \cite{HM01B0HimaxTechnologies}. Our feasibility analysis suggests that the cost, size and energy requirements of imagers are no longer barriers to their adoption. 

What we are proposing, adopting imagers as default sensors for activity recognition, is not to reject the use of accelerometers in tangential problems such as those assessing the level or quality of activities. It is also not to dismiss a multi-modality scenario, although we view the multitude of common low-dimensional sensors (e.g. magnetometer, gyroscope) as having only a supplementary role supporting the rich information collected by imagers. We acknowledge privacy concerns in processing images---but the imagers we envisioned will not be functioning as a typical camera where images are taken for record or pleasure, but as a visual sensor whose sole function is to sense activities; as a result, any images captured will only be processed locally on device. 

It is our belief that imagers are the best place to focus our research energy in order to achieve significant progress in HAR---towards recognizing the full spectrum of human activities. Our findings demonstrate that imagers offer a far superior source of information than accelerometer sensors for activity recognition, and currently the algorithmic and hardware advancements are in place to make their adoption feasible. If we place our research focus on imagers, we anticipate further enhancement that could lead to their widespread deployment and a vast array of sensing opportunities in HAR. We hope our work will lay the foundation for subsequent research exploring image-based HAR. 

\begin{figure}[t!]
		\centering
		\includegraphics[width=0.9\linewidth]{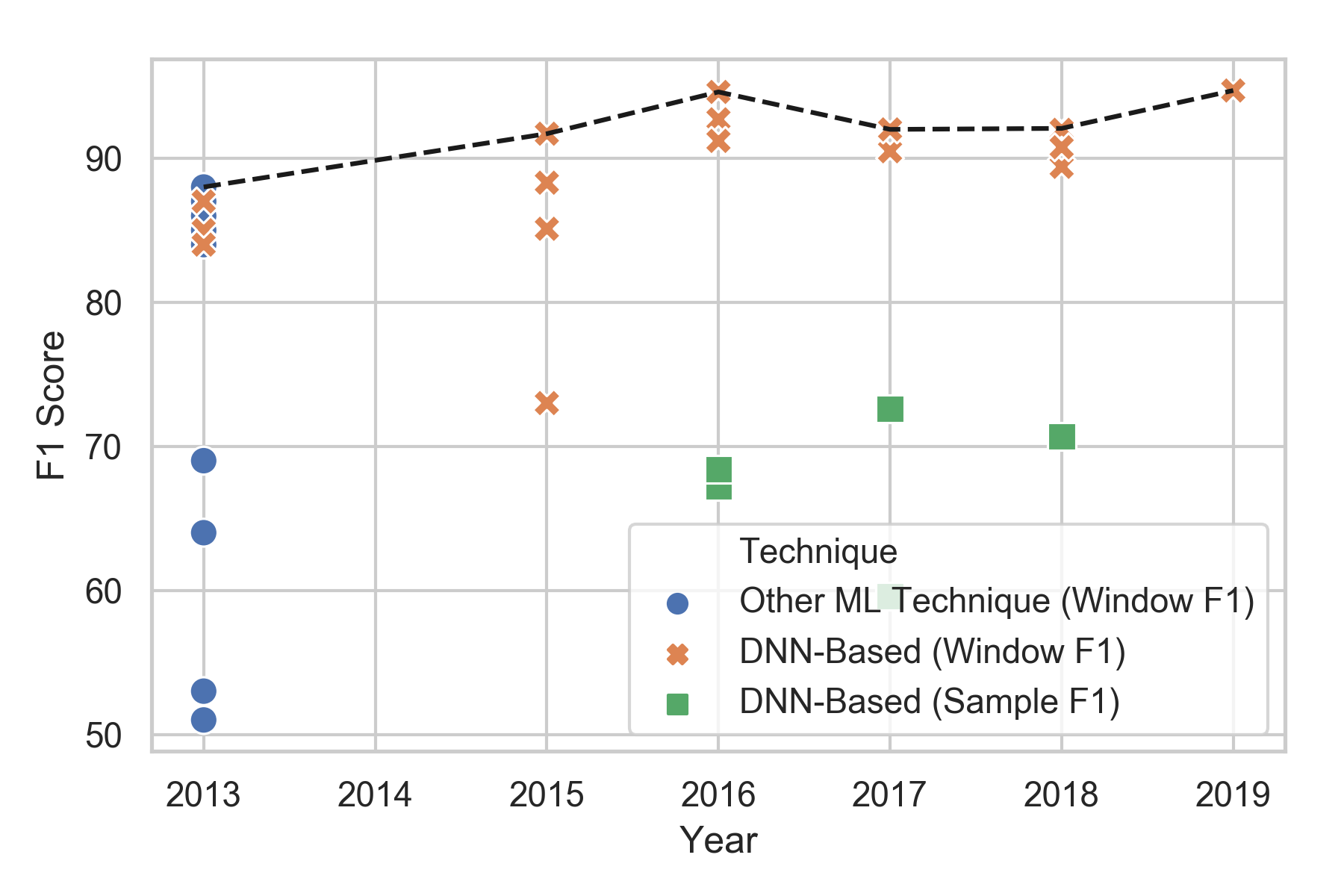}
		\caption{\small Opportunity gesture recognition, up to two scores plotted per publication per metric. We plot both the window-based F1 metric, and the sample-based F1 metric, which more recent publications prefer. We observe that deep learning has not yielded significant improvements to F1 score, despite its obvious success in other fields. Sample-based F1, believed to be a more accurate proxy to performance, is still not good enough to accurately disambiguate different activities that a user might perform.}
		\label{fig:sub1}
\vspace{-1em}
\end{figure}

\section{How Far Have We Come?}

We begin our case for imager-based activity recognition in place of acceleration-based sensing by considering where our attention to accelerometers has got us in activity recognition. In the remaining discussion, we refer to the problem of HAR exclusively as the recognition of activities from sensor data through the use of machine learning models. We review accelerometer-based HAR, summarise current challenges and finally examine how progress in HAR has stalled as a result.

\vspace{+0.5em}

\noindent \textbf{A Brief History.} {} Accelerometers respond to the intensity and frequency of motions, allowing them to provide motion-based features especially useful for characterising repetitive motions with a short time window \cite{bulling2014tutorial}. Activities such as sitting, standing, walking can be recognized efficiently using accelerometer data. The use of accelerometer data to recognize human activities have spanned almost two decades, with many initial works focused on recognising ambulation and posture \cite{mantyjarvi2001recognizing}. In 2004, Bao and Intille used multiple accelerometers to recognise 20 activities in a naturalistic setting, extending to daily activities such as watching TV and folding laundry \cite{Bao2004ActivityRF}. Since then, numerous accelerometer-based activity recognition works have followed \cite{cornacchia2016survey}, and accelerometers are perhaps the most frequently used body-worn sensor in HAR. The ubiquity of accelerometer-based activity recognition works is further cemented by the inclusion of accelerometers in almost all publicly-available datasets \cite{roggen2010collecting, skoda, pamap2}, which are used as benchmarks and thus define the scope of baseline activity recognition tasks. Beyond activity recognition, accelerometers have been used in applications including monitoring of physical activities, breathing, disease rehabilitation \cite{patel2012review}, driving pattern analysis~\cite{johnson2011driving} and skill assessment~\cite{khan2015beyond}. 

\vspace{+0.5em}

\noindent \textbf{Common Struggles.} {} 
With their motion-based features, accelerometers are most useful for identifying activities with characteristic motions, such as walking patterns. However, even within the regime of elementary actions, accelerometer-based sensing often struggles to overcome key challenges such as individual and environmental variations, sensor diversity and sensor placement issues, to which many common activities are prone \cite{lane2011community}. In turn, additional resources, often in the form of collecting more data or developing more complex models, are required to overcome the confusion between activities or to provide person-specific training. 

A fundamental challenge arises when accelerometers are used to recognise more complex activities without distinctive motions: Accelerometers inherently lack the dimension to caption any important contextual and social information beyond motion. Thus, the recognition of the full spectrum of human activities is unattainable with accelerometers.

Another widely recognised challenge with accelerometers is the difficulty of labelling the data collected, which prohibits the development of large datasets useful for deep learning techniques. Unless there is a clear ground truth was obtained closed to data collection (by recollection or additional visual data source), it is nearly impossible to label the data. 

\vspace{+0.5em}

\noindent \textbf{Stalled Progress.} {} We perform the following analysis to answer our core question: Have we really made progress with accelerometer-based activity recognition or has it stalled? We review current models and datasets, in particular, the 18-Class Gesture Recognition task from the OPPORTUNITY Activity Recognition Dataset \cite{chavarriaga2013opportunity}. We make three key observations which indicate stalled progress: 
\vspace{+0.3em}

\noindent   
\textit{Accuracy is not improving.} {} Though our learning models have matured, activity recognition has seen no dramatic jump in performance. Figure~\ref{fig:sub1} shows the weighted F1-score reported for the Opportunity gesture recognition task following its publication. Despite an initial increase in modelling power brought by Deep Neural Networks (DNN), introducing increasingly complex models has only led to scattered performance improvements in the past 5 years.

\vspace{+0.3em}
\noindent \textit{Limited Activity Labels.} {} The variety of activity labels present in publicly-available HAR datasets is limited and confined to low to medium-level activities lacking any fine-grained or contextual descriptions. The labelling of the activities has perhaps been restricted to activities that motion-based models have some hope of distinguishing, which means we can never expect activity classes at the level of differentiating eating candy versus taking pills. In most datasets, a large percentage of activities belong to a `null' or `other' class which, prospectively, we have no chance of understanding what was actually performed from the acceleration traces. 
\vspace{+0.3em}

\noindent \textit{Alarming confusions.} {} Current inertial-based HAR models are far from robust and reliable. Lack-of-common-sense errors are commonplace, e.g. confusing `drink from cup' with `opening dishwasher' in Opportunity gesture recognition \cite{yang2015deep}. The performance of activity recognition measured by different metrics can also be drastically different---more recent publications consider sample-based measurements to be more relevant (shown in green in Figure~\ref{fig:sub1}) which only give the state-of-the-art F1-scores at low 70s \cite{guan2017ensembles}, not to mention that these models already use many more sensors than are realistically deployable, as Opportunity collects data from accelerometers, gyroscopes, magnetometers at multiple body locations. 

	\begin{figure}
		\centering
		\includegraphics[width=0.8\linewidth]{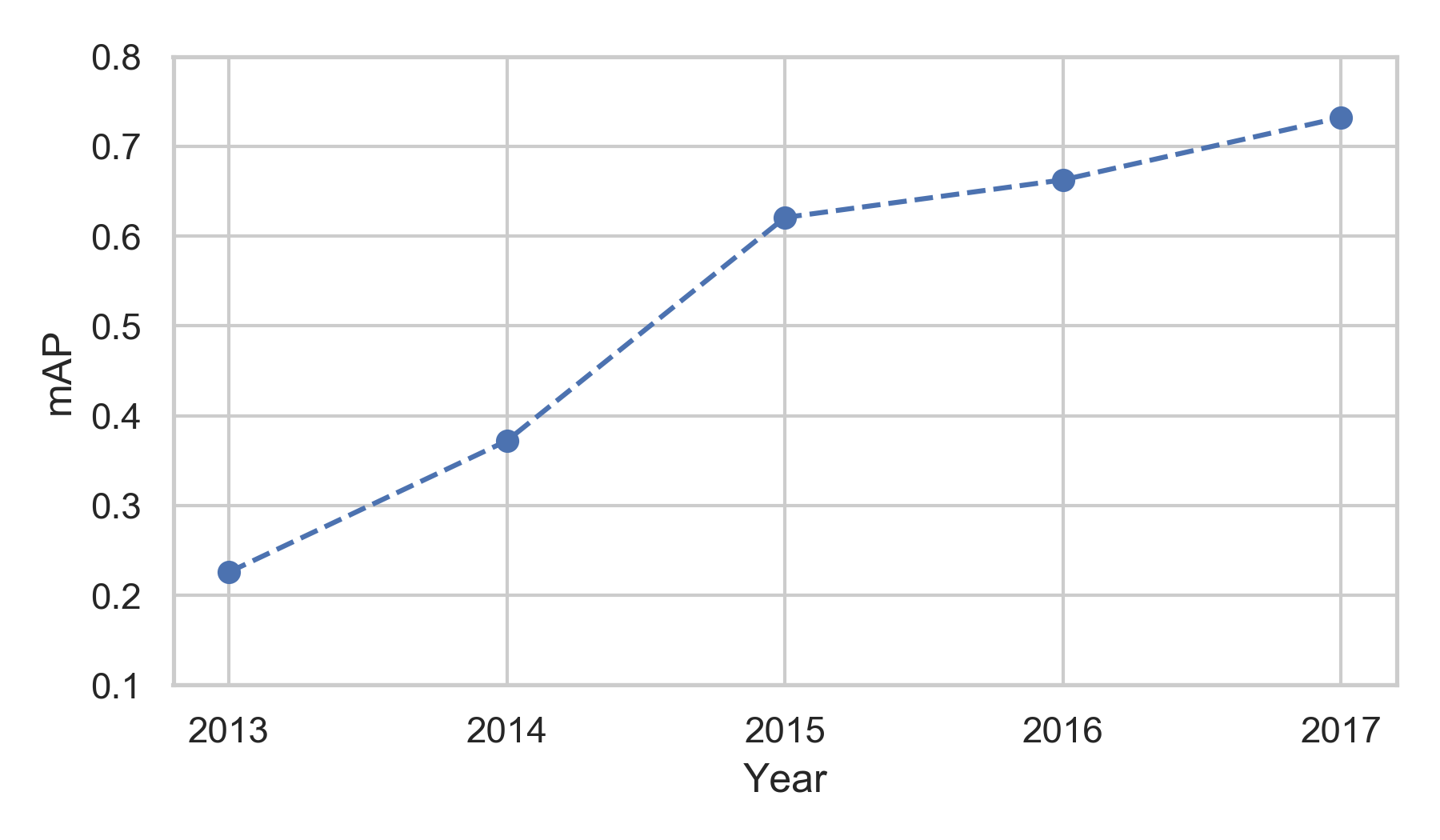}
		\caption{\small Best mAP achieved on ImageNet yearly between 2013--2017. Unlike Figure~\ref{fig:sub1}, we observe that performance for computer vision tasks---widely recognised as being challenging---have seen dramatic improvements in recent years thanks to deep learning.}
		\label{fig:sub2}
\end{figure}

\section{Are Imagers the Answer?}

It appears we have reached the end of the road for HAR with accelerometers. We believe solutions exist in sensors that capture far richer observations about activities and context. We believe that \emph{imagers}, in an embedded form which model images locally, are well-positioned to further HAR by enabling granular and contextual activity recognition. In what follows, we support our argument by pointing out that images are a rich source of information, and that the progress in computer vision provides a foundation from which we could build efficient, cheap and accurate imager-based HAR.

{\centering

\begin{figure*}
    \centering
    \includegraphics[width=0.7\linewidth]{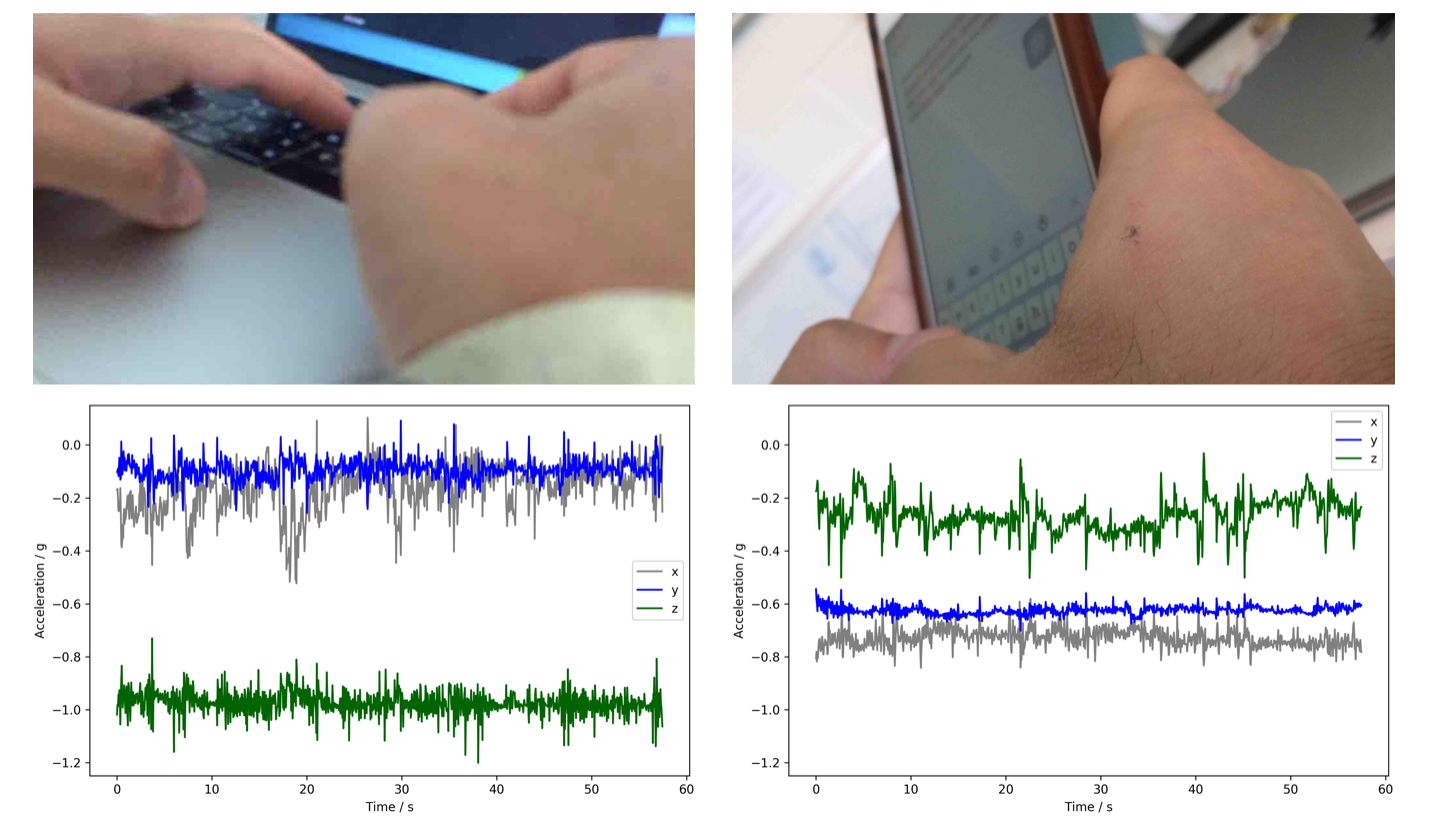}
    \caption{\small Image and accelerometer data collected from devices located at the wrist. Left: using a laptop, right: using a phone. The dominant difference in the acceleration data alignment is caused by differences in hand orientation, which can hardly be a robust feature for accurate predictions when variations within and between persons are considered. On the other hand, the differences between the two images are clear and comprehensible. A modern CNN would likely be able to disambiguate between these two images classes.}
    \label{fig:image_vs_acc}
\end{figure*}
}

\vspace{+0.5em}

\noindent
\textbf{Not the first time.} {} The idea that images are informative is not new, as supported by the vast number of activity recognition works which have used the visual modality in the form of wearable cameras, such as glasses \cite{glasses}, head \cite{fathi2012social}, wrist \cite{maekawa}, with notable wearable devices such as the SenseCam \cite{hodges2006sensecam}. Activities recognizable using images include fine-grained activities (making tea vs coffee), social interactions \cite{fathi2012social} and context. \cite{cornacchia2016survey} includes an extensive review of wearable camera-based activity recognition works. Other than wearable cameras, visual HAR based on stationary cameras has been extensively studied by the computer vision community and we refer readers to \cite{zhang2019comprehensive} for a survey in this area.

\vspace{+0.5em}
\noindent\textbf{Information-Rich Images.} {} Image data contains substantially more information than acceleration traces. Visual details captured in images may be high-level scene features quantifying the motion or contextualizing the scene, or they may be fine-grained details which specify an object that a person is interacting with \cite{cornacchia2016survey}. The rich information provided by images not only enables more granular definitions of activity classes, it also provides abundant context for open-ended study of the subjects' behaviour.

We illustrate this richness in information by conducting a simple comparison between the data captured in the two forms: images and acceleration traces. We mounted two smartphones on a subject's dominant hand, each capturing one modality; we use the app \textit{Accelerometer} \cite{accapp} for recording acceleration, and the phone's video function for taking images. Figure~\ref{fig:image_vs_acc} juxtaposes the images with the acceleration traces when two different activities were performed: typing on a phone and typing on a computer. The differences in the acceleration traces are hardly discernible nor intuitive as compared to that presented by the images. In addition, we can visually infer information such as objects, brands, environment, social setting simply from looking at these images; such additional information can be easily interrogated from the images through further processing and learning.

\vspace{+0.5em}
\noindent\textbf{Deep Learning.} {} More than ever, we have the image processing technology to be able to make use of the rich information in images, especially with the advances in Convolutional Neural Networks (CNNs). Figure~\ref{fig:sub2} shows the yearly mean average precision (mAP) reported in the ImageNet visual recognition challenge \cite{deng2009imagenet}, a benchmark in the computer visual community. There is an impressive and steady improvement in performance since the introduction of CNNs, and today the model performance has already surpassed that of human labelling; this is in stark contrast to the stalled progress seen in HAR. More than any other modality, the pairing of images and CNNs makes it practical to extract powerful information reliably enough to make the inference of complex human activities viable. In the realm of visual HAR with stationary cameras, state-of-the-art performance on the challenging (400-class) Kinetics dataset~\cite{kay2017kinetics} has achieved top-1 accuracy exceeding 60\%, although there is work to be done in leveraging these models for imager-based HAR, as will be discussed in  Section~\ref{sec:road}.

\vspace{+0.5em}
\noindent\textbf{Now is the time.} {} Finally, it is possible to accelerate CNNs substantially---to the point that on-device inference is viable, which is precisely what is needed for the required combination of images, CNNs and tiny devices. We present a detailed feasibility analysis in Section~\ref{sec:feasibility}.

\vspace{+0.5em}
\noindent\textbf{Multi-sensor?} {} We do not reject a multi-sensor scenario, especially if ample resources are available in terms of energy and computational costs. However, we view the multitude of common low-dimensional sensors (e.g. magnetometer, gyroscope) as having only a supplementary role supporting the rich information collected by imagers. Imagers remain the best choice amongst common inertial and non-inertial sensors (including microphones) due to the rich information they provide that can be used for disambiguating activities and avoiding confusions. 

\vspace{+0.5em}
\noindent \textbf{Privacy.} {} We envision imagers to have only one function: to visually sense activities, and so any image captured would only be processed \textit{locally} on the device. Running on-device recognition models prevents private visual information from ever leaving the local device end to the cloud. Concurrently, techniques developed in extreme-low-resolution activity recognition \cite{ryoo2018extreme} may be incorporated to achieve privacy by using low-resolution hardware (which further reduces computational costs). 

\vspace{+0.5em}

While prior works using image-based approaches for activity recognition exist, they largely came before CNNs, as well as other efficient on-device technologies---two key pieces to the viability of image-based HAR which were previously missing, and without which it was not worth pushing for mainstream adoption of imagers for HAR. \emph{Now} is the ideal time to move away from accelerometers and adopt imagers for activity recognition, due to the maturity of these complementary components. As a community, we now have the techniques to build models that allow us to understand images at a higher level than ever before, and the ability to run these models on smaller and more efficient devices than those envisaged even 5 years ago.

\section{Is It Practical To Use Imagers?} \label{sec:feasibility}

Conventional wisdom suggests an image-based wearable is impractical. In this section, we show that this belief has been outdated by advances in image sensing, microprocessors and machine learning. 

\vspace{+0.5em}

\noindent
\textbf{Image Sensing Technology.} {} The current power consumption associated with collecting images is sufficiently low that it does not represent a bottleneck. Over the past decade, both academia and industry have contributed to a substantial reduction in the power consumption of image sensors, and it is on the trajectory towards increasing efficiency by another order of magnitude~\cite{hansonSubMicrowattCMOSImage2010a}. 

Low form-factor imagers can now be purchased easily off-the-shelf. One such imager, capable of capturing grayscale QQVGA images at 30 frames per second (FPS), can do so using less than 1mW~\cite{HM01B0HimaxTechnologies}. We can expect a lower frame rate for activity recognition as there is no need to capture images at a frequency as high as 30FPS, hence power consumption can be reduced even further.

\vspace{+0.5em}

\noindent
\textbf{On-Device Image Recognition.} {} On-device modelling is vital to imager-based HAR: not only is it more efficient, it is also a solution to privacy concerns. A number of techniques have been proposed recently to allow neural networks to run on resource-constrained platforms such as wearable devices \cite{sze2017efficient, han2015deep, lane2017squeezing}. 
With techniques such as quantization and depthwise-separable convolutions it is now possible to run CNNs such as MobileNet~\cite{howardMobileNetsEfficientConvolutional2017} which obtain acceptable accuracy on images from ImageNet on ARM Cortex-M hardware, with sufficiently low latency for real-time use. We also expect the inference performance on low-power microcontrollers to improve rapidly: specialised RISC-V devices targeting this market are already available~\cite{KendryteKendryte}, and ARM have announced several features that will vastly improve their next generation of microcontrollers in this area~\cite{armArmEnablesCustom}.

\begin{table}[t]
\begin{tabular}{@{}lr@{}}
\toprule
\textbf{Quantity}                                    & \textbf{Value}    \\ \midrule
Latency of MobileNet ({\small$\alpha=0.25$, 160$\times$160 input}) & 165ms             \\
Microcontroller and camera run power                            & 170mW             \\
Microcontroller sleep power                          & 2mW               \\
Battery capacity                                     & 0.56Ah            \\
Safety factor                                        & 0.7               \\
Runtime                                              & \textbf{13 hours} \\ \bottomrule
\end{tabular}
\caption{\small Summary of battery life calculation using an STM32H7 microcontroller and a 3.7V 150mAh battery, which represent realistic hardware choices for a wearable device.}
\label{tab:battery_life}
\vspace{-1em}
\end{table}

\noindent \textbf{Estimating Power Consumption. }
We provide an estimate of the expected battery life using existing hardware in Table~\ref{tab:battery_life}. In our calculations, we consider the cost of deploying a MobileNet model onto a high-performance microcontroller (STM32H7), which proceed as follows: According to \cite{capotondiEEESlabMobilenetV12019}, it takes 165ms to run a MobileNet with a width multiplier 0.25 and input shape 160$\times$160$\times$3. The microcontroller consumes $\sim$170mW when running at maximum frequency, and under 2mW during sleep. This comes to an average power consumption of $\sim$30mW if processing at 1 fps. We also assume a realistic battery for a wearable device would be a 150mAh 3.7V LiPo. With a safety factor of 0.7 to account for further losses, we obtain a battery life of \textit{13 hours of continuous monitoring}, including the cost of capturing images.

Since the considered microcontroller is not designated as a low-power model, the power consumption may be assumed to reduce further using lower power manufacturing nodes. Also, the power consumption is dominated by the inference latency, which is likely to decrease over time with the integration of special-purpose accelerators or new instruction set extensions. However, even without these factors, we still obtain a useful battery lifetime, which could be extended by duty cycling readings where appropriate.

\vspace{+0.5em}
\noindent \textbf{Image Acquisition} {}
Many statistical and learning-based methods exist to overcome low-light level and blurring issues. Using other imaging modalities (e.g. IR, depth) could also alleviate these. One may also install multiple imagers at different viewpoints, though additional strains on resources are expected due to increases in model size, memory pressure and inference time. A simple concatenation pipeline for fusing images with CNNs would mean inference time scaling linearly with the number of cameras used. 

\vspace{+0.5em}

\begin{figure}[t]
	\centering
	\includegraphics[width=0.8\linewidth]{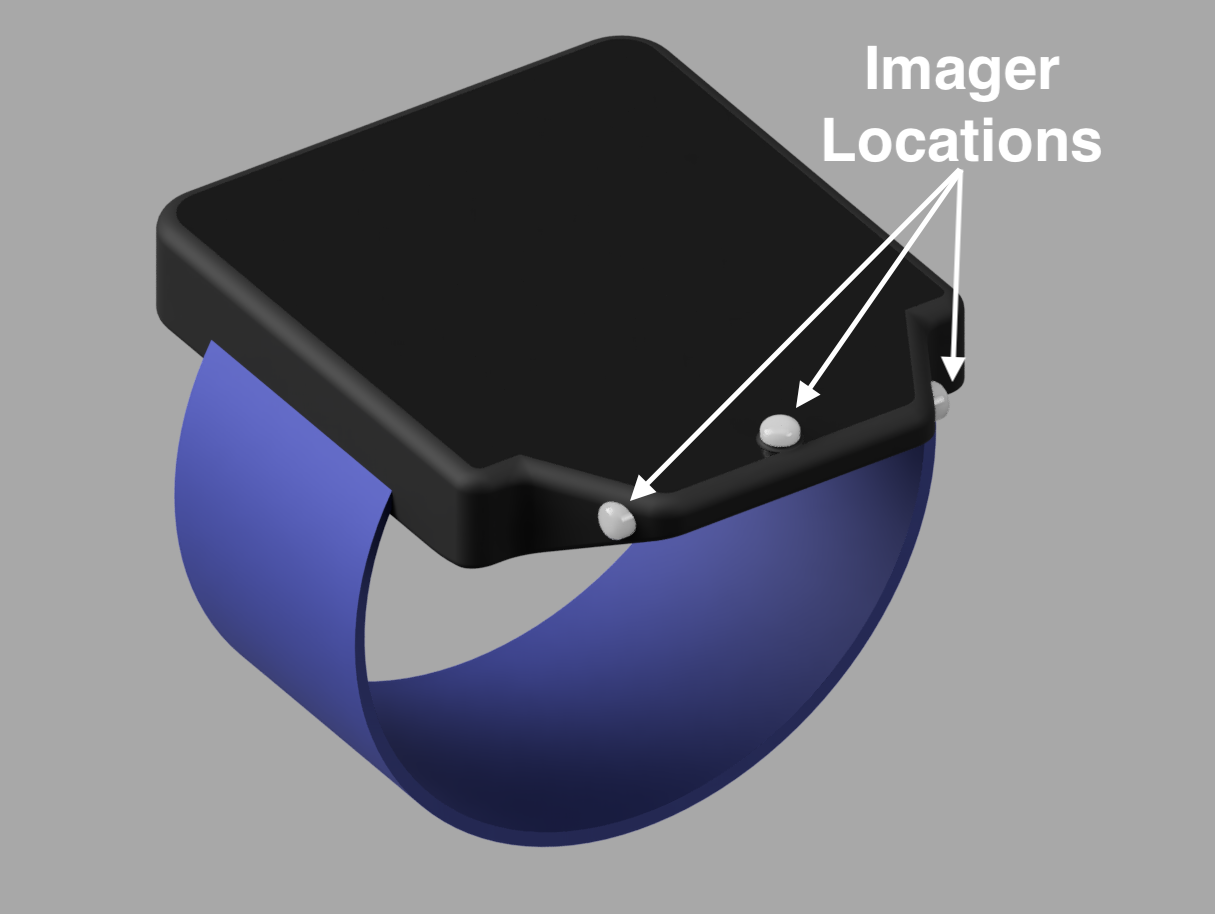}
	\caption{\small Conceptual image of a wrist-worn device with multiple embedded image sensors facing different directions. This device incorporates three sensors in a small protrusion along one side of the device in order to alleviate the problem of occlusion by clothing. This device is one of many potential form factors for imager-based HAR, which in this case takes a similar style to the common smartwatch.}
	\label{fig:wrist}
\end{figure}

\noindent
\textbf{Form Factor.} {} We believe imagers would be best placed in a form factor that is wearable and socially-acceptable while allowing for unobstructed capture of images. The choice of form factor is closely related to the orientation of imagers, the availability of continuous image streams as well as lighting conditions. Glasses are a good candidate as they can provide an unobstructed view of the wearer's current focus, which might be difficult in other cases for which specific design considerations are needed. The smartwatch is a popular form of body-worn device that imagers could be integrated into; here, the major challenge would be occlusion by clothing; this issue can be mitigated by carefully designing the device so that the sensors are placed as far down the arm as possible, along with using fish-eye lenses to increase field of view; Figure~\ref{fig:wrist} shows an artist's conception of one such device, which is one of many potential form factors for imager-based HAR.

\vspace{+0.5em}

\noindent
\textbf{Not All Scenarios.} {} While our discussion of imagers for activity recognition has mainly existed in the context of the core HAR task of identifying activities, we acknowledge that there are certain domains adjacent to activity recognition for which imagers might be harder to use or perform worse than accelerometers. In particular, imagers may fall short for sensing applications looking to measure the physical level of activities (e.g. detecting tremble levels in freezing of gait episodes of Parkinson's patients \cite{6107762}), or to assess the quality of activities (e.g. skill level \cite{khan2015beyond}). 

\section{The Road Ahead} \label{sec:road}
In this section, we present a research agenda which will make imager-based activity recognition viable. 

\vspace{+0.5em}
\noindent \textbf{Modelling.} {} We discuss key issues in modelling imagers.

\vspace{+0.3em}
\noindent \textit{From images to activities.} {} Imager-based HAR methodologies are not extensively studied, though there are many shared similar challenges with existing fields, such as egocentric image processing, odometry and object recognition. A key problem is to effectively model sparse snapshots of activities which also have a temporal relationship. An approach, given the sophistication of visual object recognition, is a multi-step approach: recognize objects then activities \cite{philipose2004inferring}. Much can also be learnt from visual action recognition methods (e.g. \cite{wang2015action}), although activities of interest in HAR might differ, and these models typically consume high-frame-rate videos not likely to be available from imager-embedded devices.

\vspace{+0.3em}
\noindent \textit{Training.} {} A core challenge is the lack of imager-generated datasets annotated for activities, especially from various ego-centric viewpoints from the body (e.g. wrist, belt, foot). Large image datasets used in object recognition (e.g. ImageNet \cite{deng2009imagenet}) or in visual action recognition (e.g. Kinetics \cite{kay2017kinetics}) present potential opportunities for transfer learning from these respective domains to that of naturalistic egcocentric images; such transfer learning schemes will be highly fruitful by leveraging non-sensitive, standardized datasets.

\vspace{+0.3em}
\noindent \textit{Multi-views.} {} To effectively combine data generated from multiple imagers embedded on a device, one might leverage multi-camera fusion from robotics \cite{keyes2006camera}, though accomplishing this from an egocentric perspective with sporadic actions is a challenge.

\vspace{+0.3em}

\noindent \textit{Community efforts.} {} The availability of accessible imager datasets is vital to facilitating modelling efforts in imager-based HAR. The ideal scenario is the development of such a dataset, with the community also defining a wide scope of activity labels, so that models can be built to recognize activities of different complexity at different stages of development. Doing so while addressing privacy concerns of collecting such datasets would be another open challenge.

\vspace{+0.5em}
\noindent \textbf{Image Sensors.} {}
Our feasibility analysis had been run with off-the-shelf image sensors sub-optimal for imager-based HAR. The camera selected also provides better specifications than necessary: both the frame rate and image size were larger than what current hardware can process on-device.
It will be possible to build cameras that more closely match the target application, with lower power consumption and manufacturing costs, e.g. adopting voltage reduction ~\cite{hansonSubMicrowattCMOSImage2010a}.

\vspace{+0.5em}
\noindent \textbf{Hardware Design.} {}
Finally, machine learning on low-power devices is an area that is receiving significant attention from both the academic and industrial community.
Microcontrollers incorporating extensions or accelerators suited for machine learning workloads are already becoming a reality~\cite{KendryteKendryte, armArmEnablesCustom}, which would cause a significant reduction in inference latency.
There is concurrent work on how to train networks that utilise this hardware optimally.

\section{Conclusion}
Accelerometers are a dead-end for activity recognition: they do not offer enough information and our reliance on them has led to a stalled progress in HAR. In order to recognize the full spectrum of human activities we should adopt imagers as the default HAR sensor, since it is now possible to exploit the information-richness of images with the rise of energy-efficient CNNs. Collectively, our study argues that now is the time to pursue HAR using imagers, and calls for further exploration into overcoming the existing challenges for imager-based HAR.

\section*{Acknowledgement}
This work was supported by the \grantsponsor{EPSRC}{Engineering and Physical Sciences Research Council (EPSRC)}{} under Grant No.: DTP (\grantnum[]{EPSRC}{EP/R513295/1}) and MOA (\grantnum[]{EPSRC}{EP/S001530/}), and \grantsponsor{samsung}{Samsung AI}{}. We would like to thank Dr.~Petteri Nurmi and other anonymous reviewers for their feedback throughout the submission process for ACM HotMobile 2020, and William Ip for facilitating the experiments.

\input{chopped-definitely-final-hopefully.bbl}


\end{document}